\documentclass[runningheads]{llncs}
\usepackage[T1]{fontenc}
\usepackage{cite}
\usepackage{verbatim}
\usepackage{amsmath,amssymb,amsfonts}
\usepackage{comment}
\usepackage{algorithmic}
\usepackage{graphicx}
\usepackage{textcomp}
\usepackage{orcidlink}
\usepackage{marvosym}
\usepackage{pdfpages}
\usepackage{multicol}
\usepackage{booktabs}
\usepackage{multirow}
\usepackage{bigstrut}
\usepackage{xspace}
\usepackage{xcolor}
\usepackage{xspace}
\usepackage{graphicx}
\usepackage{anyfontsize}
\usepackage{hyperref}

\begin{document}
\title{Concept Alignment Contrast and Long-Short Prompt Memory for Test-Time Adaptation of SAM3 in Medical Image Segmentation
}
\titlerunning{Concept Alignment Contrast and Long-Short Prompt Memory}
%

%
%
\author{Yubo Zhou\inst{1}  \and Jianghao Wu\inst{2} \and Ping Ye\inst{1} \and Shaoting~Zhang\inst{1,3} \and Guotai Wang\inst{1,3}}

\authorrunning{Y. Zhou et al.}

\institute{
University of Electronic Science and Technology of China, Chengdu, China  \and Faculty of Information Technology, Monash University, Melbourne, VIC 3800, Australia \and Shanghai Artificial Intelligence Laboratory, Shanghai, China \email{guotai.wang@uestc.edu.cn}}

%
\maketitle              
%
\begin{abstract}

%
Concept segmentation models like Segment Anything Model 3 (SAM3) show strong generalization on natural images, yet their performance degrades in medical imaging due to the domain gap caused by different imaging principles and styles.
%
Test-Time Adaptation (TTA) is essential for improving the testing performance by updating the model on the fly without annotations. However, existing vision-language TTA methods are mainly driven by image-level uncertainty minimization, which does not necessarily reflect region-level semantic correctness in medical segmentation. Moreover, they often lack mechanisms to maintain stability in continual one-pass adaptation, leading to limited performance when reliable dense supervision is missing for segmentation.
To address these issues, we propose \textbf{C}oncept Alignment Contrast and Long-Short Prompt \textbf{M}emory for \textbf{T}est-\textbf{T}ime \textbf{A}daptation (CM-TTA) of SAM3 for medical images. First, for a test sample with multiple augmentations, we introduce a novel Concept Alignment Contrast (CAC) metric, which leverages textual-visual semantic consistency to robustly evaluate prediction quality and select the best augmented view for pseudo-label generation.
Second, to balance rapid and stable adaptation, we design a Long-Short Prompt Memory (LSPM) module. The short memory dynamically fuses recent prompts based on CAC scores for agile local adaptation, while the long memory maintains a stable global prompt to generate enhanced pseudo-labels. Finally, a Densely Supervised Prompt Update (DSPU) strategy is proposed to optimize the prompt embeddings with enhanced pseudo labels as dense supervision. Extensive experiments on prostate and skin lesion segmentation demonstrate that our CM-TTA framework significantly outperforms existing methods for TTA of SAM3. The code is available at \url{https://github.com/SherlockZYB/CM-TTA}.

\end{abstract}
\section{Introduction}
Medical image segmentation plays an important role in clinical applications~\cite {sharma2010automated}. Traditional deep learning models have achieved impressive performance, but struggle to generalize to unseen imaging modalities and segmentation targets~\cite{wang2022medical}. Recently, vision-language models such as the Segment Anything Model 3 (SAM3)~\cite{carion2025sam} have achieved high generalizability to unseen object classes due to the textual prompt-driven segmentation paradigm with zero-shot inference ability. However, it still struggles in medical imaging due to substantial semantic and visual domain gaps. Although recent efforts~\cite{liu2025medsam3,jiang2026medicalsam3} fine-tune SAM3 on large medical datasets, such retraining with annotations is costly. Some unsupervised adaptation methods~\cite{xu2023novel,liu2023memory} can adapt the model to a target domain without human annotations, but they need a large batch of data, which is infeasible for an online stream of testing samples, limiting their real-world deployment.

To tackle this problem, Test-Time Adaptation (TTA) has emerged as a practical solution without annotations~\cite{liang2025comprehensive, sheng2025illusion}. Traditional TTA methods~\cite{he2021autoencoder,wang2021tent,wang2022cotta,li2024cache,liu2022single} focus on the visual modality of pre-trained segmentation models, updating models based on auxiliary tasks~\cite{he2021autoencoder,liu2022single}, entropy minimization~\cite{wang2021tent}, or pseudo-labels~\cite{wang2022cotta,li2024cache} for adaptability. However, they fail to leverage the rich textual modality that is important for providing semantic guidance to ensure consistency during cross-domain adaptation. With the rise of vision-language models, text-driven TTA~\cite{shu2022tpt,lu2024histpt,xiao2025dynaprompt,imam2025ttl} paradigms have been increasingly explored. For instance, TPT~\cite{shu2022tpt} optimizes prompts on the fly by minimizing entropy across multiple augmented views, 
while TTL~\cite{imam2025ttl} updates low-rank adapters by confidence maximization. Despite their improvements, they heavily rely on image-level uncertainty metrics (e.g., entropy~\cite{shu2022tpt} or difference~\cite{xiao2025dynaprompt}) as a criterion to select the reference prediction or optimization object, which is insufficient for guiding accurate pixel-level predictions in medical segmentation. Besides, these methods only track the instantaneously updated parameters, which are sensitive to the current test batch, and hardly maintain stability in long-term adaptation.

To address these limitations, we propose \textbf{C}oncept Alignment Contrast and Long-Short Prompt \textbf{M}emory for \textbf{T}est-\textbf{T}ime \textbf{A}daptation (CM-TTA) of SAM3 for medical image segmentation. Our method adapts SAM3 to a new domain efficiently by prompt tuning~\cite{shu2022tpt} through three key components. First, we introduce a Concept Alignment Contrast (CAC) metric to evaluate the prediction quality. For a test image with multiple augmented views, CAC computes the contrast between foreground-text and background-text similarities to explicitly leverage text-driven semantic guidance, selecting the optimal augmented view for subsequent adaptation. Second, we design a Long-short Prompt Memory (LSPM) module, where the short prompt memory stores recent prompts and corresponding CAC scores, which is used to generate a fused prompt for better prediction output. While the long prompt memory is updated by the momentum of the short prompt memory to maintain historical representations for stable adaptation. Third, we propose a Densely Supervised Prompt Update (DSPU) strategy, where enhanced pseudo-labels obtained by the long-term prompt and the CAC-selected view are used to supervise predictions from views with the short-term prompt. Extensive experiments on Promise12~\cite{litjens2014promise} and ISIC2018~\cite{codella2019isic} demonstrate that CM-TTA significantly improves SAM3's test-time performance on prostate and skin lesion segmentation tasks by an average Dice gain of 3.17 and 6.07 percentage points over the baseline.

\section{Method}

\begin{figure*}[t]
	\centering
	\centerline{\includegraphics[width=11cm]{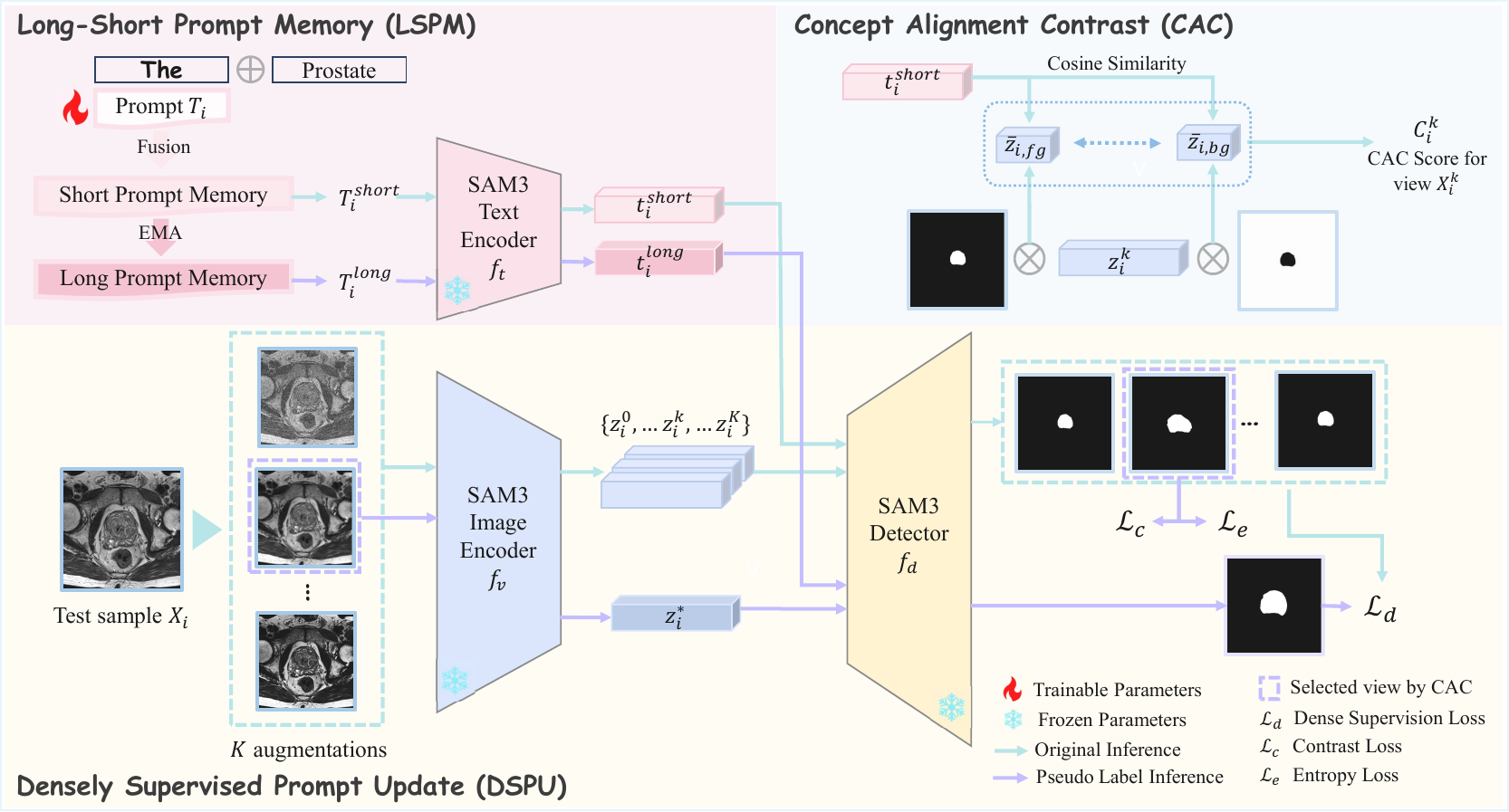}}
   \caption{\textcolor{black}{Overview of our CM-TTA for test-time adaptation of SAM3.}
  } \label{fig:pipeline}
\end{figure*}
Previous work has shown that directly fine-tuning all parameters of foundation models is computationally expensive and can reduce the generalizability,  while prompt tuning avoids distorting pre-trained representations and enables lightweight adaptation~\cite{kumar2022fine,wortsman2022robust,shu2022tpt}. 
Therefore, we adopt prompt tuning for SAM3 by updating only the learnable text prompt token embeddings $T\in\mathbb{R}^{N\times D}$. 
$T$ is initialized from a tokenized phrase (e.g., ``the prostate''), where $N$ and $D$ denote the number of prompt tokens and the embedding dimension, respectively.
Given a SAM3 model pre-trained on natural images with a frozen text encoder $f_t$, vision encoder $f_v$, and detector $f_d$, the goal is to adapt SAM3 to an unseen medical target domain $\mathcal{D}_\mathcal{T}=\{X_i\}_{i=1}^{|\mathcal{D}_\mathcal{T}|}$ that exhibits distribution shifts from SAM3's pre-training data. 
At each time step $i$, we perform a one-step prompt update on $X_i$ based on CM-TTA, and then generate the final segmentation for $X_i$ with the updated prompt before proceeding to $X_{i+1}$. 
This yields a one-pass continual TTA protocol, where each test sample is used exactly once for adaptation~\cite{wang2022cotta,chen2024each}.

As illustrated in Fig.~\ref{fig:pipeline}, our CM-TTA consists of three components: 
(1) Concept Alignment Contrast (CAC) evaluates text--visual semantic alignment between predicted foreground/background regions and the prompt across multiple augmented views, and selects the most reliable view for guiding the other views; 
(2) Long-Short Prompt Memory (LSPM) maintains a short-term prompt bank for rapid local adaptation and a momentum-updated long-term prompt as a stable semantic anchor; and 
(3) Densely Supervised Prompt Update (DSPU) generates enhanced pseudo-labels using the long-term prompt on the CAC-selected view and uses them as dense supervision to update $T$.

\subsubsection{Concept Alignment Contrast (CAC)}
Prediction quality estimation is crucial for TTA to avoid confirmation bias or error accumulation without ground truth~\cite{lee2024aetta,shu2022tpt,lu2024histpt}, but existing estimation methods like entropy~\cite{shu2022tpt} or variance~\cite{wu2024fpl+} are mainly designed for visual modality, which are not well calibrated with the textual information. We therefore propose CAC, a text-guided quality score that measures the degree to which the prompt is more aligned to the predicted foreground than the background, indicating the prediction confidence. 


Let $T_i$ denote the prompt state before adapting on the $i$-th test image $X_i$. 
Given $X_i$ and the pre-update prompt $T_i$, we extract the visual feature $z_i=f_v(X_i)$ and the textual feature $t_i=f_t(T_i)$, and feed them into the detector to obtain a soft prediction $P_i=f_d(t_i,z_i), P_i\in[0,1]^{H\times W}$.
We then obtain the foreground and background visual evidence by masked average pooling:

\begin{equation}
    \bar{z}_{i,{fg}} = \frac{\sum_{u} P_i(u)\, z_i(u)}{\sum_{u} P_i(u) + \epsilon}, \quad
    \bar{z}_{i,{bg}} = \frac{\sum_{u} (1-P_i(u))\, z_i(u)}{\sum_{u} (1-P_i(u)) + \epsilon},
\end{equation}
where $u$ indexes spatial locations and $\epsilon$ is a small constant. Subsequently, the CAC score is defined as
\begin{equation}
    C_i = \mathrm{sim}\!\left(\bar{z}_{i,{fg}}, t_i\right)
    - \mathrm{sim}\!\left(\bar{z}_{i,{bg}}, t_i\right),
\end{equation}\label{eq:cac}
where $\mathrm{sim}(\cdot,\cdot)$ denotes the cosine similarity. By encouraging foreground--text alignment while suppressing background--text similarity, CAC provides a reliable criterion for view selection under distribution shift. Given original view $X_i^0=X_i$ and $K$ augmented views $\{X_i^k\}_{k=0}^{K}$, the CAC score of each view $C_i^k$ can be calculated with visual feature $z_i^k=f_v(X_i^k)$, and the view with the highest score $C_i^*$ will be selected as the optimal view $X_i^*$.

\subsubsection{Long-Short Prompt Memory (LSPM)}
\label{sec:lspm}
Continual one-pass TTA methods mostly track the instantaneously updated parameters only, making them sensitive to the current sample, and hardly maintain stability in long-term adaptation~\cite{wang2022cotta,xiao2025dynaprompt}. To achieve a balance between immediate adaptation and long-term stability, we introduce LSPM that combines a short-term prompt for rapid adaptation with a momentum-updated long-term prompt for semantic stabilization. 

Short Prompt Memory (SPM): We maintain a first-in-first-out memory $\mathcal{M}_i=\{(T_j,C_j^*)\}_{j=\max(1,\,i-L)}^{i-1}$ of length $L$ that stores the most recent prompts and their optimal CAC scores. To use the recent knowledge for rapid adaptation and emphasize historically reliable prompts, we use CAC score as weights fuse historical prompts in $\mathcal{M}_i$:
\begin{equation}
    T_i^{hist}=\sum_{(T_j,C_j^*)\in\mathcal{M}_i} w_j T_j,
    \qquad
    w_j=\frac{\exp(C_j^*)}{\sum_{m=\max(1,\,i-L)}^{i-1}\exp(C_m^*)}.
\end{equation}

Then, to balance the impact of historical prompts and encourage efficient integration of recent historical knowledge, we calculate the CAC scores of $T_i^{hist}$ and $T_i$ with $X_i$ to get $C_i^{hist}$ and $C_i$ according to Eq.~\ref{eq:cac}, respectively. Next, a dynamic weight 
$w_{hist}= \frac{\exp(C_i^{hist})}{\exp(C_i^{hist})+\exp(C_i)}$ is used to get a short memory prompt instead of $T_i$ for the original forward propagation with multiple views:
\begin{equation}
    T_i^{{short}}=w_{hist}T_i^{{hist}}+(1-w_{hist})T_i.
\end{equation}
For optimal view selection in CM-TTA, Eq.~\ref{eq:cac} is then instantiated with $T_i^{short}$.

Long Prompt Memory (LPM): To improve stability, we maintain a long-term global prompt updated by Exponential Moving Average (EMA):
\begin{equation}
    T_i^{{long}}=\gamma T_{i-1}^{{long}}+(1-\gamma)T_i^{{short}},
\end{equation}
where $\gamma\in[0,1)$ is the momentum. 
$T_i^{{long}}$ changes smoothly over time and provides a stable representation for generating enhanced pseudo-labels.

\subsubsection{Densely Supervised Prompt Update (DSPU)}
\label{sec:dspu}
Dense pixel-level supervision is crucial for accurate medical segmentation, yet ground-truth masks are unavailable at test time~\cite{mishra2022data}. 
We therefore propose DSPU, which uses an enhanced pseudo-label generated by the long-term prompt $T_i^{{long}}$ on the CAC-selected view $X_i^*$ to provide dense self-supervision. We compute the visual feature $z_i^*=f_v(X_i^*)$ and detached text embedding $t_i^{{long}}=f_t(T_i^{{long}})$, and obtain a soft pseudo-label:
\begin{equation}
    P_i' = f_d\!\left(t_i^{{long}},\, z_i^*\right), \qquad P_i'\in[0,1]^{H\times W}.
\end{equation}

Using $P_i'$ as the reference, we encourage the short-term prompt-based prediction in all $K+1$ views to be consistent with it by a soft Dice loss:
\begin{equation}
\label{eq:ld}
    \mathcal{L}_{d}
    = 1-\frac{1}{K+1}\sum_{k=0}^{K}
    \frac{2\sum_{n=1}^{H\times W} P_i'(n)\,P_i^k(n)}
    {\sum_{n=1}^{H\times W} P_i'(n)+\sum_{n=1}^{H\times W} P_i^k(n)+\epsilon},
\end{equation}
where $P_i^k=f_d\!\left(t_i^{{short}},\, z_i^k\right)$ is the soft prediction for view $X_i^k$, $n$ indexes pixels.

We further maximize the CAC score via $\mathcal{L}_c=-C_i^*$ to encourage confident textual-visual semantic alignment, and also apply pixel-wise entropy minimization loss $\mathcal{L}_{e}=-\frac{1}{H\times W}\sum_{n=1}^{H\times W}\Big[P_i^*(n)\log P_i^*(n)+(1-P_i^*(n))\log(1-P_i^*(n))\Big]$ to encourage confident predictions. As applying the confident maximization and entropy minimization regularization to low-quality predictions may reinforce prediction errors, we only apply $L_c$ and $L_e$ to the optimal view $P_i^*$ selected by $C_i^*$. The total loss for adaptation is
\begin{equation}
\label{eq:ltotal}
    \mathcal{L}=\mathcal{L}_{d}+\alpha \mathcal{L}_{c}+\beta \mathcal{L}_{e},
\end{equation}
where $\alpha$ and $\beta$ are loss weights. 
For each sample $X_i$, we perform a single back-propagation to get the updated prompt $T_{i+1}$, and then output the final segmentation $P_{final}=f_d(f_t(T_{i+1}),f_v(X_i))$ using the updated prompt.

\section{Experiments}
\begin{table}[htbp]
\centering
\caption{Comparison between different TTA methods on the Promise12 and ISIC2018 Datasets using SAM3 as the source model. \dag denotes a significant improvement (p-value < 0.05) over the best existing method.}
\label{tab:quantitative_results}
\resizebox{0.95\textwidth}{!}{
\begin{tabular}{lccccc}
\toprule
\multirow{2}{*}{Method} & \multirow{2}{*}{\begin{tabular}[c]{@{}c@{}}Trainable\\Params (K)\end{tabular}} & \multicolumn{2}{c}{Promise12} & \multicolumn{2}{c}{ISIC2018} \\
\cmidrule(lr){3-4} \cmidrule(lr){5-6}
 & & Dice (\%) $\uparrow$ & ASSD (pixel) $\downarrow$ & Dice (\%) $\uparrow$ & ASSD (pixel) $\downarrow$ \\
\midrule
No adapt & 0.00 & 70.79 $\pm$ 26.59 & 14.59 $\pm$ 20.76 & 76.27 $\pm$ 26.22 & 22.45 $\pm$ 27.30 \\
ZERO~\cite{farina2024zero} & 0.00 & 71.17 $\pm$ 26.37 & 13.75 $\pm$ 18.92 & 80.10 $\pm$ 22.33 & 17.82 $\pm$ 23.51 \\
TPT~\cite{shu2022tpt} & 1.02 & 72.05 $\pm$ 26.49 & 12.57 $\pm$ 19.79 & 79.60 $\pm$ 23.33 & 18.08 $\pm$ 24.89 \\
HisTPT~\cite{lu2024histpt} & 1.02 & 71.15 $\pm$ 27.11 & 13.73 $\pm$ 20.97 & 78.95 $\pm$ 24.70 & 18.51 $\pm$ 27.86 \\
TENT~\cite{wang2021tent} & 163.84 & 71.19 $\pm$ 26.81 & 13.86 $\pm$ 20.56 & 76.45 $\pm$ 26.50 & 21.94 $\pm$ 27.82 \\
TTL~\cite{imam2025ttl} & 655.36 & 71.60 $\pm$ 27.26 & 13.08 $\pm$ 20.84 & 75.61 $\pm$ 28.14 & 20.82 $\pm$ 31.15 \\
Ours & 1.02 & \textbf{73.96 $\pm$ 24.36}$^\dag$ & \textbf{10.91 $\pm$ 17.71}$^\dag$ & \textbf{82.34 $\pm$ 20.42}$^\dag$ & \textbf{9.39 $\pm$ 14.38}$^\dag$ \\
\bottomrule
\end{tabular}
}
\end{table}
\subsubsection{Datasets and Implementation Details}
In the experiments, we adapted SAM3 to two medical image segmentation datasets:  1) Promise12~\cite{litjens2014promise} that consists of 100 transverse T2-weighted Magnetic Resonance (MR) volumes collected from four medical centers, 1475 2D slices with foreground were used for experiments; 2) ISIC2018~\cite{codella2019isic} that contains 2954 images of various types of skin lesions at different resolutions. For preprocessing, the Promise12 slices were resized to $320\times320$, while the ISIC2018 images were resized to $256\times256$.

For both datasets, the image intensity was normalized to the range of [0,1]. Each test sample undergoes a one-step update with the Adam optimizer and a learning rate of 0.005. The text prompt is set as `the prostate' for Promise12, and `the skin lesion' for ISIC2018 with corresponding 1 token (i.e., $N=1$) of dimension $D=1024$ being optimized. For CM-TTA, the randomly applied augmentation includes Gaussian blur, color jitter, contrast/sharpness adjustment, and we set $K=9$. The $\gamma$ was set to 0.99 for the EMA according to previous work~\cite{tarvainen2017mean}, and $\beta$ was set to 0.1 for the entropy loss to avoid over-confidence. Contrast loss weight $\alpha$ and short prompt memory bank length $L$ were set to 1.0 and 16 according to the ablation study.



\subsubsection{Comparison with State-of-the-art TTA Methods}

Our CM-TTA was compared with five methods: 
1) \textbf{TENT}~\cite{wang2021tent} that updates parameters of normalization layers of the vision encoder by entropy minimization; 2) \textbf{TPT}~\cite{shu2022tpt} that selects the low-entropy views in the augmented batch and optimizes the textual prompt;
3) \textbf{TTL}~\cite{imam2025ttl} that proposes to optimize low-rank module inside the vision encoder to minimize a weighted entropy; 4) \textbf{ZERO}~\cite{farina2024zero} that directly conducts a vote on low-entropy augmented views without training; 5) \textbf{HisTPT}~\cite{lu2024histpt} that updates prompt with historical prompt memories. `No adapt' serves as the baseline, meaning inference with SAM3 without adaptation. Method-specific hyperparameters of the compared methods were tuned for optimal results. 

\begin{figure*}[t]
	\centering
	\centerline{\includegraphics[width=11cm]{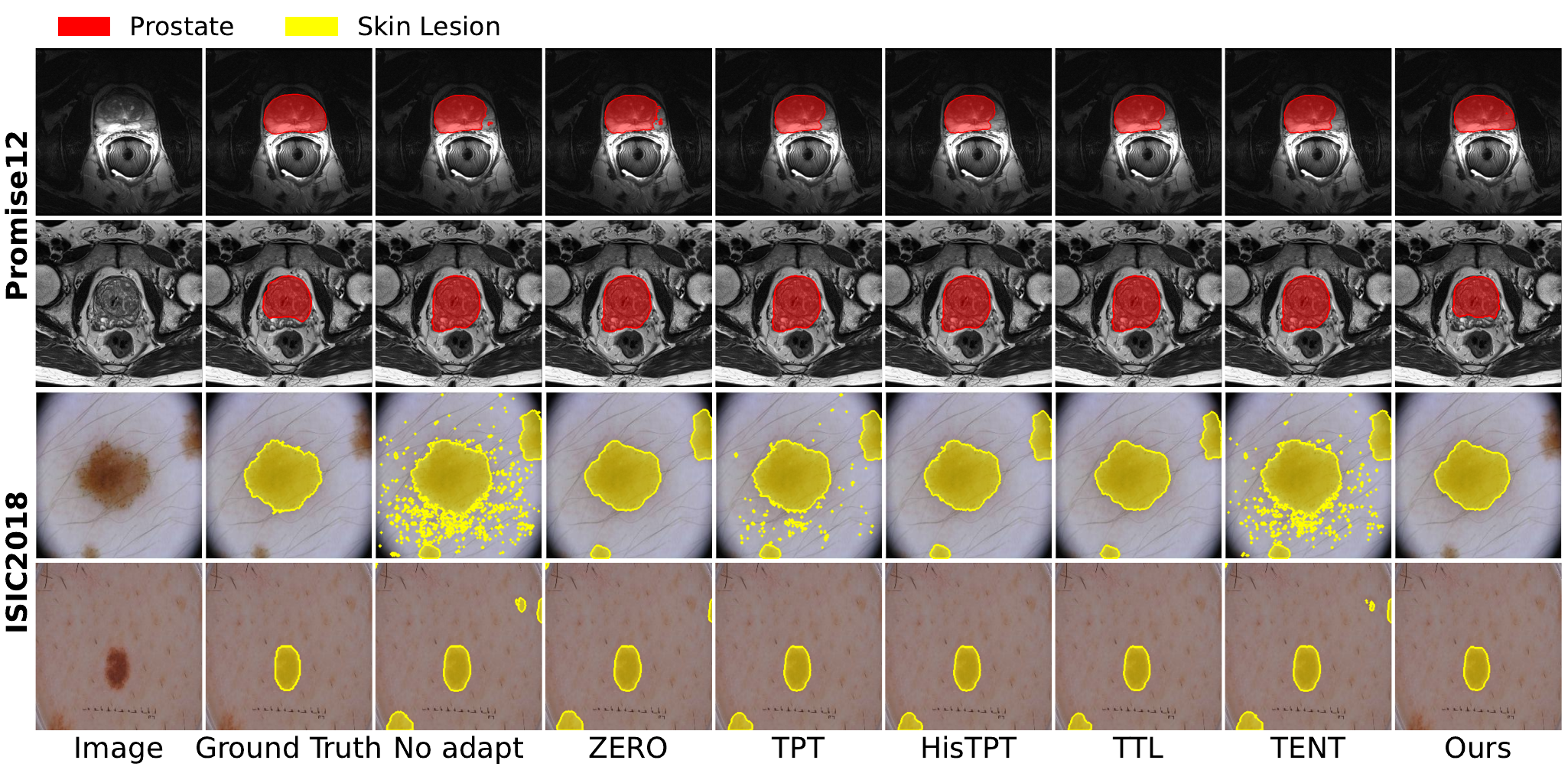}}
   \caption{\textcolor{black}{Qualitative comparison of different TTA methods on two datasets. }
  } \label{fig:segmentation}
\end{figure*}
\begin{table}
\centering
\caption{Ablation study of CM-TTA. 
Row 2 means using CAC for selecting the best augmented view for direct inference without model update, while rows 3-6 mean using the pseudo-labels from the CAC-selected view for supervising the other views in DSPU. DSPU=$\circ$ means using $\mathcal{L}_d$ only, and DSPU=$\bullet$ means using $\mathcal{L}_d$+$\mathcal{L}_c$.}
\label{tab:ablation_study}
\resizebox{0.9\textwidth}{!}{
\begin{tabular}{cccccccc}
\toprule
\multirow{2}{*}{CAC} & \multicolumn{2}{c}{LSPM} & \multirow{2}{*}{DSPU} & \multicolumn{2}{c}{Promise12} & \multicolumn{2}{c}{ISIC 2018} \\
\cmidrule(lr){2-3} \cmidrule(lr){5-6} \cmidrule(lr){7-8}
 & LPM & SPM & & Dice (\%) $\uparrow$ & ASSD (pixel) $\downarrow$ & Dice (\%) $\uparrow$ & ASSD (pixel) $\downarrow$ \\
\midrule
 & & & & 70.79 $\pm$ 26.59 & 14.59 $\pm$ 20.76 & 76.27 $\pm$ 26.22 & 22.45 $\pm$ 27.30 \\
$\checkmark$ & & & & 71.79 $\pm$ 25.35 & 13.89 $\pm$ 19.37 & 78.47 $\pm$ 24.63 & 19.30 $\pm$ 27.01 \\
$\checkmark$ & & &$\circ$ & 71.90 $\pm$ 26.61 & 12.82 $\pm$ 19.68 & 78.82 $\pm$ 22.85 & 18.78 $\pm$ 26.60 \\
$\checkmark$ & & &$\bullet$ & 72.53 $\pm$ 26.17 & 13.10 $\pm$ 19.87 & 78.96 $\pm$ 24.44 & 18.93 $\pm$ 26.80 \\
$\checkmark$ & & &$\checkmark$ & 72.90 $\pm$ 24.85 & 13.09 $\pm$ 17.32 & 79.10 $\pm$ 24.24 & 17.80 $\pm$ 24.09 \\
$\checkmark$ & $\checkmark$ & & $\checkmark$ & 73.39 $\pm$ 24.04 & 11.38 $\pm$ 17.02 & 79.19 $\pm$ 23.84 & 11.06 $\pm$ 14.98 \\
$\checkmark$ & $\checkmark$ & $\checkmark$ & $\checkmark$ & \textbf{73.96 $\pm$ 24.36} & \textbf{10.91 $\pm$ 17.71} & \textbf{82.34 $\pm$ 20.42} & \textbf{9.39 $\pm$ 14.38}\\
\bottomrule
\end{tabular}
}
\end{table}
As shown in Table~\ref{tab:quantitative_results}, the unadapted SAM3 achieved the worst results with an average Dice of 70.79\% and 76.27\% on Promise12 and ISIC2018, respectively. Existing TTA methods yield moderate improvements over the baseline; the best existing method on Promise12 is TPT~\cite{shu2022tpt} with an average Dice of 72.05\% and an average ASSD of 12.57 pixels, while on ISIC2018, ZERO~\cite{farina2024zero} obtained an average Dice of 80.10\% and an average ASSD of 17.82 pixels. In contrast, our CM-TTA significantly outperforms all the existing methods on both datasets with only 1.02K trainable parameters, which is 160 and 650 times fewer than TENT~\cite{wang2021tent} and TTL~\cite{imam2025ttl}, respectively. We achieve the highest Dice scores of 73.96\% on Promise12 and 82.34\% on ISIC2018, and reduce the ASSD to 10.91 pixels and 9.39 pixels, respectively. Besides, CM-TTA only takes an average of 0.57s for the inference of one test sample on Promise12, which is close to the inference time using unadapted SAM3 directly (0.31s).

The qualitative results shown in Fig.~\ref{fig:segmentation} demonstrate that existing methods often result in under-segmentation or over-segmentation, while our CM-TTA is superior in accurately delineating the target regions in different domains with variation of contrast (prostate) and object size (skin lesion).

\subsubsection{Ablation Study}

Quantitative results of the ablation study on two datasets are shown in Table~\ref{tab:ablation_study}. It can be observed on both datasets that using CAC to select the optimal view as the final results without updating can directly improve the performance. Then, by introducing DSPU with multiple loss step by step, the performance is further improved. Finally, the incorporation of LSPM contributed to the best performance, indicating the effectiveness of combining LPM for robust semantic anchors and SPM for efficient short-term adaptation. 

\begin{figure}[t]
	\centering
	\centerline{\includegraphics[width=11.5cm]{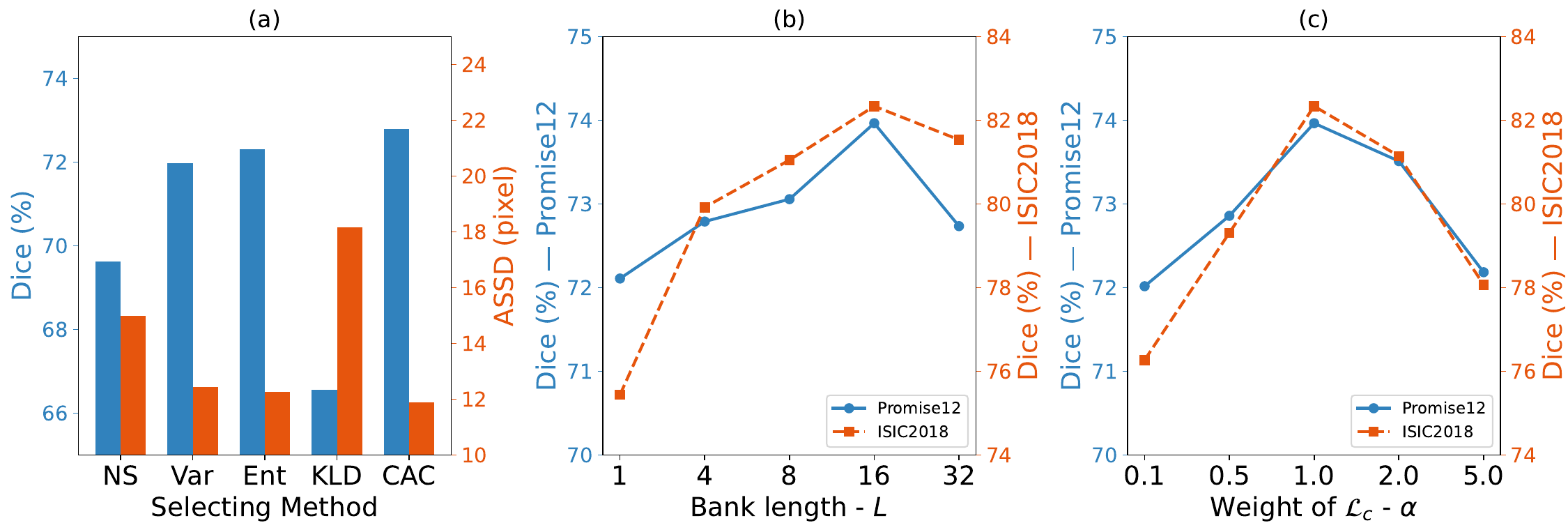}}
   \caption{(a) Comparison of the quality of the best augmented view selected by different metrics on Promise12. (b) Effect of short memory bank length $L$ on average Dice. (c) Effect of the weight of $\mathcal{L}_c$ on average Dice.
  } \label{fig:hypa}
\end{figure}
Besides, for the metric to select the augmented view on testing images, we compare our CAC with some alternative methods, including Entropy~\cite{shu2022tpt}, Variance~\cite{wu2024fpl+}, Kullback-Leibler Divergence (KLD) with mean prediction~\cite{yang2022dltta}, and No Selection (NS) that uses the mean predictions as the result.  Fig.~\ref{fig:hypa} (a) shows that CAC outperformed other methods, obtaining the highest Dice and lowest ASSD. In addition, our CM-TTA has two main hyper-parameters, i.e., the short memory bank length $L$ and the weight $\alpha$ of $\mathcal{L}_c$. Fig.~\ref{fig:hypa} (b) and (c) indicate that $L=16$ and $\alpha=1.0$ performs best for CM-TTA, and they are generalizable for different datasets, making it easily applied to new datasets.

\section{Conclusion}
We proposed Concept Alignment Contrast and Long-Short Prompt Memory for Test-time Adaptation (CM-TTA) of SAM3 in medical image segmentation. Our work addresses the severe over-confidence and confirmation bias inherent in conventional uncertainty-based TTA methods when applying text-driven segmentation models to a new medical image segmentation task. By introducing CAC, we effectively leverage multi-modal semantic consistency to evaluate prediction quality without requiring ground truth. Furthermore, the proposed LSPM successfully navigates the plasticity-stability dilemma in continuous TTA, ensuring agile continuous learning for local adaptation while maintaining stability. Combined with the DSPU, our framework establishes an efficient, self-correcting adaptation loop. Extensive experiments on two datasets demonstrate that CM-TTA achieves state-of-the-art performance. In future work, it is of great clinical interest to extend this framework to 3D volumetric medical segmentation.


\subsubsection{Disclosure of Interests.}
The authors have no competing interests to declare that are relevant to the content of this article.
\bibliographystyle{splncs04}
\bibliography{paper-2125}
\end{document}